# Synthetic linguistic data

Giuseppe Samo (Beijing Language and Culture University) samo@blcu.edu.cn

**Abstract**

Synthetic linguistic data have become increasingly prominent in linguistic research, from the analysis of large collections of AI-generated texts to controlled experiments investigating grammatical and lexical preferences. This glossary entry extends Bresnan's (2016) distinction between the linguistic garden and the bush by introducing an artificial plant nursery as a metaphor for synthetic data. We distinguish between synthetic bush data, consisting primarily of corpus-like collections of model-generated language, and synthetic garden data, in which model outputs are elicited under controlled conditions to investigate specific linguistic hypotheses. We argue that synthetic data constitute a distinct object of linguistic investigation: comparing them with natural language data can reveal both shared patterns and divergences, but their interpretation requires careful consideration of the conditions under which they are generated.

## 1 Introduction

The role of experimental methods and the use of corpora have long been debated in the history of linguistics, particularly with respect to the relationship between controlled linguistic judgements and naturally occurring language data (Chomsky 1956; Gibson and Fedorenko 2013; Sprouse and Almeida 2013). Bresnan's (2016) distinction between the garden and the bush provides a powerful metaphor for capturing this methodological contrast. The garden represents carefully controlled data: sentences and minimal pairs tested through introspective or large-scale judgements, but also curated corpora and grammatical examples designed to describe specific phenomena or test specific hypotheses. By contrast, the bush refers to naturally occurring language data, as found in corpora (Gries 2008). This glossary entry argues that the current landscape of linguistic data requires an extension of Bresnan's metaphor to include a third space, an artificial plant nursery, dedicated to synthetic data.

Synthetic and simulated data in general have long been integral to data science and predictive modelling (Nobani et al. 2026 and references therein). They exist with respect to linguistic data, although their boundaries are considerably more complex to define (Samo & Merlo 2021, 2026a). A first type of synthetic data typically associated with the term is the corpus-like output of ConversationalAI based on large language models (LLMs) (De Cesare 2026), which we present in section 2. However, synthetic data also include other forms of model-generated material obtained under carefully controlled conditions. Examples include data produced through masked language modelling and fill-mask procedures, which have also been widely used to augment curated datasets for large-scale evaluation, as illustrated by benchmarks such as BLiMP (Warstadt et al. 2020) and Blackbird

Language Matrices’ datasets (e.g. An et al. 2023; Jiang & Merlo 2026). Further examples are discussed in Section 3.

In both cases, these types of linguistic data are neither straightforwardly garden data nor direct samples from the bush. Instead, they are produced in partly opaque environments. Like plants in a nursery, they may resemble natural specimens, while exhibiting properties that distinguish them from both naturally occurring and experimentally elicited data. Building on the distinction between the garden and the bush, this glossary entry examines two complementary roles of synthetic data in linguistic inquiry: (i) synthetic bush data and (ii) synthetic garden data.

Before turning to the definitions, an important question is whether synthetic linguistic data, and the outputs of large language models in particular, can provide evidence for theories of language. Piantadosi (2023) argues that the success of LLMs may suggest that these models may constitute viable theories of language. Chesi (2025), while acknowledging their strong empirical performance, questions whether predictive success alone is sufficient to establish explanatory adequacy. Other positions argue that synthetic data may not meaningfully contribute to the study of human language as a cognitive system (Katzir 2023, Moro et al. 2023, Shi et al. 2025). These latter positions raise a broader methodological question concerning the relationship between the empirical evidence from language models and the evidence required to formulate and evaluate linguistic theories. In this glossary entry, we treat natural data and synthetic data as distinct objects of investigation, building on Rizzi (2025). This distinction allows us to examine what each type of data can reveal about language without assuming that evidence obtained from one can be reduced to the other.

Finally, synthetic data can be used to evaluate computational models through large-scale benchmarks, such as the CALAMITA challenge for Italian (Nissim et al. 2026), which draws on different types of data to assess model capabilities. Synthetic data can also be used for training, raising a different set of questions, including whether generated examples reinforce existing biases or limitations (Zhang and Pavlick 2025, Samo & Merlo 2026a).

## 2 Synthetic bush data

One of the most visible forms of synthetic linguistic data is the output of conversational AI systems based on LLMs, which has brought the Turing test (Turing 1950) and Searle’s Chinese room argument (Searle 1980) back into both academic and public debate. Outputs of ConversationalAIs can be collected into corpus-like datasets in order to investigate the properties of generated language (Cegin et al. 2026). Each model can be treated as an informant or, more generally, as a population of speakers. The relationship between these outputs and naturally occurring language data, however, remains methodologically complex. Although generated texts can be assembled into collections resembling conventional corpora, they are produced in response to prompts and under specific interactional conditions. In this respect, they can also be understood as elicited data (Di Caro & Samo 2025 and reference therein), whose properties depend not only on the model itself but also on the prompts and other generation parameters (e.g. temperature).

Mikros (2025), for instance, investigates GPT-4o's ability to imitate the literary styles of Ernest Hemingway and Mary Shelley. By comparing model-generated texts with the authors' original works using several stylometric methods and three prompting strategies, the study shows that GPT-4o can reproduce some surface-level stylistic features. Although in-context learning improves stylistic alignment, the generated texts retain similarities to more generic outputs generated by the model.

The comparison between human and machine-generated languages has become a central concern in both academic research and public debate. An excellent example is the summary in De Cesare (2026) that examines the linguistic, textual, pragmatic, and stylistic properties of *synthetic* Italian, identifying features that distinguish AI-generated texts from naturally produced Italian (De Cesare 2023, 2024). In particular, the pivotal role of English in current large language models (Wendler et al. 2025) leaves traces of English fingerprints in synthetic Italian texts (see De Cesare 2024). Moreover, prompting instructions in English can affect the characteristics of outputs generated in Italian (Cicero 2023, Nastase et al. 2025).

The use of generated outputs as linguistic evidence is not restricted to the description of synthetic language itself. Since the emergence of conversational LLMs, theoretical linguists have increasingly used them to investigate questions traditionally addressed through corpus analysis, experimental elicitation, or grammaticality judgements. Haider (2023), for example, examines ChatGPT's grammatical competence through its responses to questions concerning English and German syntax. Merlo (2023) investigated the capacity of ChatGPT to generalise over linguistic patterns using a task designed to test rule-based generalisation (e.g. subject-verb agreement). Although the model could follow the instructions and identify correspondences between objects and attributes, the study of its output shows that it failed to infer the underlying structure of the task and consistently produced incorrect solutions, even with considerable prompting assistance.

Similarly, Massaro & Samo (2023) investigated ChatGPT's metalinguistic awareness of word-order reorderings in Italian and larger or less resourced- Italian varieties. The authors found that ChatGPT tended to label topicalisation structures, for example *Il libro, Mario lo legge* 'the book, Mario reads it' as archaic or antiquated, despite evidence that these constructions occur in contemporary Italian (for example, around one fifth of direct objects are fronted in data from Italian extracted from annotated treebanks). The study also identified difficulties in interpreting object clitics (e.g. *lo* in *Mario lo legge*) and suggested that more resourced languages may overshadow minor varieties (e.g. Neapolitan and the variety spoken in San Marco in Lamis, Apulia) in the model's linguistic representations.

More broadly synthetic outputs have become an object of systematic evaluation across a wide range of tasks and domains. Beyond linguistic phenomena, research has examined models' ability to represent real-world patterns or scientific knowledge (Wysocka et al. 2024), or detect for cultural artifacts (Ghaboura et al. 2025). A related issue is sycophancy, the tendency of conversational AI systems to agree with or validate users rather than critically assess their claims. Cheng et al. (2026), for example, show that this tendency extends to situations in which users describe harmful or questionable behaviour, and that sycophantic responses can influence users' subsequent judgments. Similarly, synthetic outputs can be investigated to explore prototypical semantics. For example, Samo & Ursini (2024) shows that the semantic profile of the Italian geonym *piazza* 'square' in synthetic data differs from the patterns observed in natural-language sources, including general-purpose corpora and specialised sources.

These evaluations serve different purposes: generated outputs may be assessed to probe the capabilities and limitations of the models themselves, or the results may justify the creation of dedicated datasets to redirect models (Samo et al. 2022) so that linguists can harvest more data representative of more linguistic populations.

Finally, application-oriented research can help identify the limitations of synthetic data and caution against their uncritical use in real-world settings. For example, studies on AI-generated facilitator companions for Emergency Medical Teams operating in deep-field settings (Samo & Parker 2025, to appear) examine the use of synthetic conversational outputs in crisis response. In Samo & Parker (to appear), for instance, disaster scenarios generated by DeepSeek-V3.1 and ChatGPT-5 were evaluated in English, Italian, Chinese and Tagalog. These outputs were evaluated in terms of both their linguistic properties and their plausibility and suitability for the intended application by expert practitioners in the field and native speakers of the language under investigation. While the outputs in the first three languages received good evaluations in terms of language and content, those in Tagalog were rated poorly by speakers. Interestingly, a cross-LLM-as-a-judge comparison showed that both models rated the other model's Tagalog outputs extremely high.

Summing up, synthetic bush data can be readily collected from conversational AI systems and analysed using corpus-linguistic methods. However, their apparent ease of creation should not obscure the methodological challenges they raise. To ensure that findings are interpretable and replicable, each dataset should document relevant generation conditions, including the model used, the prompt instructions (content and language), and possibly the date of generation. Systematically documenting these conditions is therefore essential to comparing results across studies and to making generalisations about synthetic language.

## 3 Synthetic garden data

The definition of synthetic garden data is more complex. In contrast to corpus-like collections of generated texts, these data are constructed under controlled conditions to investigate specific linguistic hypotheses. In this respect, some forms of synthetic data can be understood as an extension of experimental methods traditionally used in psycholinguistics.

In an early phase of language models, masked language modelling and fill-mask procedures, for instance, can be used to elicit model responses (e.g. fill-mask procedures) to carefully designed linguistic stimuli, making it possible to investigate lexical preferences and grammatical constraints (e.g. subject-verb agreement) under controlled conditions (Linzen et al. 2016, Bernardy & Lappin 2017, Gulordava et al. 2018).

The parallelism with psycholinguistic measures to investigate the internal preferences of language models. When carefully designed, synthetic data can illuminate interactions between grammatical constraints and probabilistic preferences and may reveal parallels with language-processing mechanisms: one prominent example is *surprisal*, a measure of the information-theoretic unexpectedness of a linguistic unit given its preceding context (Hale 2016 and reference therein; see also Greco et al. 2023).

Wilcox et al. (2024) have explored how surprisal-based measures can be used to investigate the grammatical expectations encoded by neural language models and to compare model behaviour with human processing patterns in complex English constructions, including parasitic gaps and other syntactic islands. Surprisal has subsequently been applied across languages and linguistic phenomena. For instance, Samo & Merlo (2024) shows that surprisal patterns associated with movement in French clefts mirror those observed in human processing, while Samo & Chen (2026) demonstrates that degrading ratings for wh-ex-situ configurations in Chinese are already clearly marked by small-scale transformer models.

Sentence embeddings can also be analysed to investigate how language models represent abstract linguistic structure and generalise across lexical variation. For example, Samo et al. 2026 examines the representations of lexicalised sentences and their functional counterparts, in which pronouns and adverbs replace more specific nouns or phrases. The results show that the sentences with only pronominal elements and sentences with NPs occupy different regions of the embedding space, and that their shared syntactic structure does not emerge when either type is considered in isolation.

Tokenisation strategies provide a further perspective on the limitations of linguistic generalisation. Samo & Merlo (2026b) have shown that subword segmentation can affect the representation of morphological forms in complex verbal paradigms in Turkish and Hebrew, particularly when language-specific morphological patterns do not align with token boundaries. Similarly, Hopton et al.

(2025) investigate how subword tokenisation interacts with the representation of the functional lexicon across languages.

Another approach consists in directly querying conversational AI systems for numerical grammaticality/acceptability judgments. In principle, this amounts to evaluating a numerical output, closer to an experimental procedure, rather than a textual output, as in synthetic bush data. For example, Samo & Massaro (2025) investigates syntactic configurations in Romansh and Neapolitan by prompting ChatGPT and DeepSeek to rate sentences on a numerical scale. The results reveal asymmetries across models and constructions: while the predictions concerning violations to Verb Second phenomena in Romansh are partially reflected in the conversational AI ratings, topic–subject agreement in Neapolitan remains challenging. For the Romance phenomena investigated, smaller monolingual transformer models even performed better than conversational AI systems. On the other hand, related work has explored similar approaches for German, comparing LLM-generated acceptability judgments with those provided by human participants (Catasso 2026). These findings illustrate both the potential and limitations of numerical judgments elicited from conversational AI as a tool for investigating grammatical preferences across languages with different structural properties and levels of representation in language-model training data.

Synthetic garden data allow researchers to test specific linguistic hypotheses by asking models to evaluate carefully constructed sentences. Their strength lies in the possibility of comparing model responses with human judgments and the predictions of different linguistic theories using replicable methods. They can also provide a starting point for in-depth research on human language (Samo & Chen 2026).

## 4. Conclusions

In this glossary article, we introduced the distinction between natural and synthetic linguistic data, focusing on the different forms that synthetic data can take and the methodological implications of their use. The distinction between synthetic bushes and synthetic gardens highlights two complementary approaches: the collection and analysis of corpus-like model outputs, and the controlled generation of linguistic data to investigate specific linguistic phenomena.

Synthetic data offer new opportunities for linguistic research. At the same time, the interpretation of these data requires careful consideration of the conditions under which they are generated: in particular, similarities between model outputs and naturally occurring language, or between model responses and human judgements, should not automatically be taken as evidence that the underlying mechanisms are equivalent.

Rather than treating natural and synthetic data as competing sources of evidence, we follow Rizzi (2025) in considering them as distinct but potentially complementary objects of investigation. Their comparison can help identify patterns shared by human language use and model behaviour, as well as divergences that require further explanation. Future research should therefore ask not only whether synthetic data mirror natural language, but also whether they can open up new questions for linguistic research.